\documentclass{article}
\usepackage[preprint]{neurips_2024}
\usepackage[utf8]{inputenc}
\usepackage[T1]{fontenc}    
\usepackage{hyperref}       
\usepackage{url}            
\usepackage{booktabs}      
\usepackage{amsfonts}      
\usepackage{nicefrac}       
\usepackage{microtype}     
\usepackage{xcolor}         
\usepackage[pdftex]{graphicx}
\usepackage{amsmath}
\usepackage{tabularx}
\usepackage{booktabs} 
\usepackage{multirow}
\usepackage{authblk}
\usepackage{subcaption}
\usepackage{wrapfig}

\title{D2Vformer: A Flexible Time Series Prediction Model Based on Time Position Embedding}

\author[a]{Xiaobao~Song}
\author[a]{Hao~Wang\thanks{Corresponding author: haowang@szu.edu.cn}}
\author[a]{Liwei~Deng}
\author[b]{Yuxin~He}
\author[a]{Wenming~Cao}
\author[c]{Chi-Sing~Leung}
\affil[a]{Shenzhen University, China}
\affil[b]{Shenzhen Technology University, China}
\affil[c]{City University of Hong Kong, Hong Kong, China}

\begin{document}

\maketitle

\begin{abstract}
Time position embeddings capture the positional information of time steps, often serving as auxiliary inputs to enhance the predictive capabilities of time series models. However, existing models exhibit limitations in capturing intricate time positional information and effectively utilizing these embeddings. 
To address these limitations, this paper proposes a novel model called D2Vformer.
Unlike typical prediction methods that rely on RNNs or Transformers, this approach can directly handle scenarios where the predicted sequence is not adjacent to the input sequence or where its length dynamically changes. In comparison to conventional methods, D2Vformer undoubtedly saves a significant amount of training resources. 
In D2Vformer, the Date2Vec module uses the timestamp information and feature sequences to generate time position embeddings. 
Afterward, D2Vformer introduces a new fusion block that utilizes an attention mechanism to explore the similarity in time positions between the embeddings of the input sequence and the predicted sequence, thereby generating predictions based on this similarity. 
Through extensive experiments on six datasets, we demonstrate that Date2Vec outperforms other time position embedding methods, and D2Vformer surpasses state-of-the-art methods in both fixed-length and variable-length prediction tasks.
\end{abstract}

\section{Introduction}

Time series forecasting plays a pivotal role in various real-world applications, including resource allocation\cite{energy_forecasting, solar_forecasting}, traffic state prediction\cite{traffic_congestion,traffic_forecasting_1,traffic_forecasting_2}, and weather alert\cite{wind_forecasting_1,wind_forecasting_2}.
Initially, statistical methods such as exponential smoothing\cite{ex_smoothing} and auto-regressive moving averages (ARMA)\cite{arima_1} were commonly used for time series forecasting.
With the advancement of deep learning, an increasing number of deep learning models have been proposed and applied in this field.  Convolutional Neural Network(CNN) based methods(e.g., TCN\cite{tcn}) use convolution and pooling operations to extract time-series features and aggregate time step information.  Recurrent Neural Network(RNN) based methods(e.g., DeepAR\cite{deepar}, LSTNet\cite{lstnet}) utilize recurrent networks to extract time-series features and predict sequences.  Transformer-based methods(e.g., Informer\cite{informer}, TDformer\cite{tdformer}) leverage attention mechanisms to capture the similarity relationships between time steps and make predictions using an encoder-decoder architecture.

\begin{figure}[tb]
   \centering
   \includegraphics[width=1\textwidth]{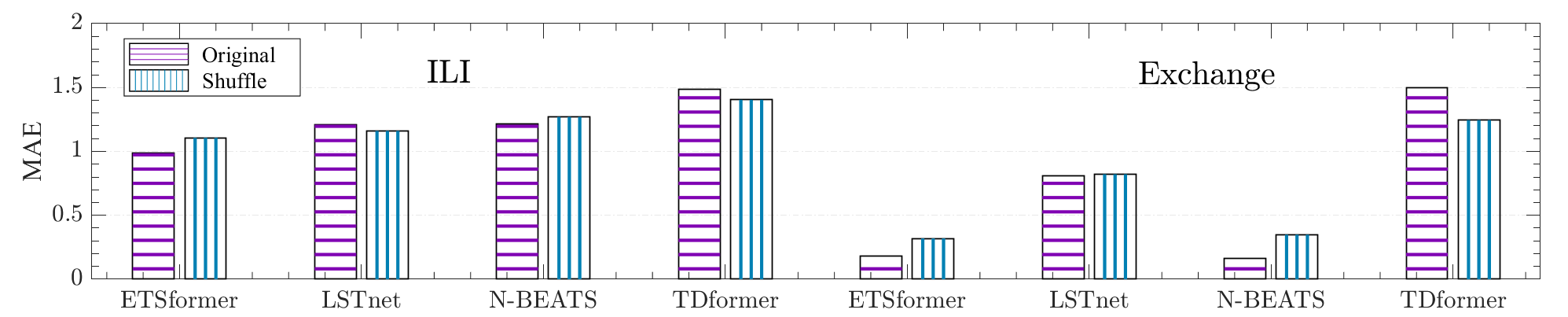}
   \caption{The model comparison experiments before and after shuffling on two datasets.}
   \label{fig:shuffle}
\end{figure}

DLinear\cite{dlinear} posits that while these models demonstrate good predictive performance in specific time series forecasting scenarios, they often lack a comprehensive exploration of time position information. 
Through empirical experiments, we have demonstrated the existence of this issue in these models.
In the experiments, we disrupt the time position information by shuffling the order of the input sequence, and then compare the models' performance before and after the disruption. 
As illustrated in Figure \ref{fig:shuffle}, we observe that most models' prediction accuracy experiences a negligible decrease after shuffling, and in some cases, even a slight increase in accuracy is observed. This finding serves as compelling evidence that these models indeed lack a comprehensive exploration of time position information. 
To address this challenge, some models incorporate time position embeddings to capture time position information. 
For instance, Transformer-based models\cite{ns_transformer,patchtst} utilize fixed-frequency sine and cosine functions to generate time position embeddings. Time2Vec (T2V)\cite{time2vec} incorporates the timestamps of time steps as input to generate time position embeddings, capturing the positional relationships among time steps through learnable linear and frequency components. For the time position embeddings generated by the aforementioned methods, existing models\cite{time2vec_app_1,time2vec_app_2,time2vec_app_3} incorporate them into the input through summation and concatenation operations.

Due to the complex and fluctuating nature of real-world prediction scenarios, the positional relationships between time steps often exhibit intricate patterns. The position embedding in Transformer, utilizing fixed-frequency sine and cosine functions, fails to capture these patterns effectively. 
Although T2V incorporates timestamps as inputs, attempting to capture the linear and periodic patterns of time positions, the periodicity information contained in timestamps is limited.
In time series analysis, periodic patterns of time positions are often reflected in feature sequences. 
Therefore, our goal is to develop a time position embedding method that leverages timestamps and feature sequences to uncover the underlying intricate patterns of time positions.
To achieve this goal, we introduce a novel module called Date2Vec (D2V) for time position embedding. This module considers both timestamps and feature sequences, offering two key advantages:
(1) D2V can leverage both timestamps and feature sequences to learn the linear positional relationships and complex periodic relationships between time steps.
(2) The time position embedding by D2V is data-driven, enabling it to adjust based on changes in the data dynamically.

Another challenge faced by existing models is their inability to make flexible predictions, specifically in generating predicted sequences of varying lengths during inference. 
Existing models typically rely on specific network layers\cite{informer,autoformer,fedformer} or the autoregressive approaches\cite{gru,lstm_1} to generate predictions. 
However, specific network layers require predefined input and output dimensions, limiting the model's ability to adjust the length of the predicted sequence. Although the autoregressive approaches allow for adjusting the prediction length during inference, they are highly likely to lead to issues like gradient explosion\cite{lstm_2,lstm_3} and error accumulation\cite{error_accumulation_1,error_accumulation_2}. To overcome this challenge, we propose a Fusion Block, which uses the batch matrix-matrix product to generate predicted sequences without the need for a predefined output sequence length, thus enabling flexible prediction.

Based on the above ideas, we propose a new time series forecasting model named D2Vformer\footnote{The code for D2Vformer is available at: \url{https://github.com/TCCofWANG/D2Vformer}.}. 

Specifically, D2Vformer mainly involves two stages: (1) D2V: generating time position embeddings that encompass intricate patterns. (2) Fusion Block: calculating the positional relationship between the input and predicted sequence through batch matrix-matrix product and generating the predicted sequence based on this relationship.
  
In summary, the main contributions of our work include:
\begin{itemize}
  \item  We propose Date2Vec, a novel time position embedding method that can effectively capture intricate positional relationships among time steps.

  \item  We propose the Fusion Block, which utilizes the time position embeddings generated by D2V and outputs the predicted sequence based on batch matrix multiplication without requiring a predefined length of the output sequence.

  \item  D2Vformer is capable of directly handling time series prediction tasks with varying prediction lengths and tasks involving intervals between input and output sequences.
  
  \item The experimental results show that D2Vformer outperforms other state-of-the-art methods in both regular and flexible prediction scenarios. Also, the Date2Vec and Fusion Block outperform their respective comparison algorithms.
\end{itemize}

\section{Background}

\begin{figure}[t]
    \centering
    \includegraphics[width=0.8\textwidth]{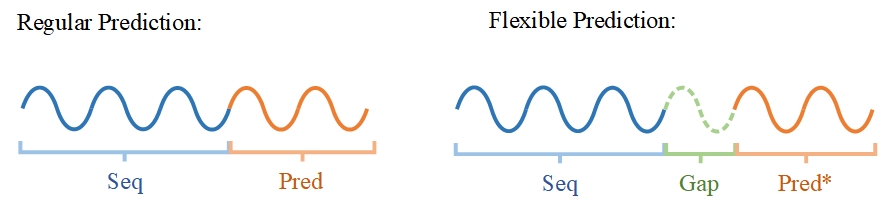}
    \caption{The left figure shows the regular prediction for time series. The right one illustrates flexible prediction. In both figures, the blue solid line represents the input sequence, the orange solid line represents the predicted sequence, and the green dashed line represents the interval between them. The predicted length ‘Pred*’ in flexible prediction is variable.}
    \label{preliminary}
\end{figure}

In time series prediction tasks, models typically extract temporal features from input historical sequences and then utilize these features to generate predicted sequences. Given an input historical time series $\pmb{X} \in \mathbb{R}^{L \times D}$ ($L$ is the length of the historical series and $D$ denotes the variable dimension) and the predicted time series $\hat{\pmb{Y}} \in \mathbb{R}^{O \times D}$ ($O$ denotes the prediction length), the regular time series prediction task can be expressed as:
\begin{equation}
\hat{\pmb{Y}} = \text{model} (f(\pmb{X})),
\end{equation}
where $f(\cdot)$ represents the feature extraction module, $\text{model}(\cdot)$ represents the prediction model. 
To improve performance, some prediction models incorporate time position embedding methods. These methods typically utilize timestamp information as input to generate embeddings capturing the positional relationships between time steps.
Taking time series prediction tasks with dates as the basic unit, time position embedding involves converting each time step into a date vector, as detailed in Appendix \ref{date_embedding}.
Thus, the time position information of the input sequence is represented as an embedding matrix $\pmb{D}_{x} \in \mathbb{R}^{L \times M}$, and similarly for the predicted sequence, $\pmb{D}_{y} \in \mathbb{R}^{O \times M}$, where $M$ is the dimension of the date vector.
These time position embeddings are usually concatenated with temporal features as inputs to the prediction model. Consequently, time series models with time position embedding methods can be described as:
\begin{equation}
\hat{\pmb{Y}} = \text{model} (Concat(f(\pmb{X}),E(\pmb{D}_{x}))),
\end{equation}
where $E(\cdot)$ represents the time position embedding method, and $Concat(\cdot,\cdot)$ is the concatenation operation. 
However, these models still face two primary challenges. First, existing time position embedding methods struggle to fully capture the complex patterns in the time series. 
Second, simply concatenating the time position embeddings with the input feature can not effectively utilize this information. To address these issues, we introduce a new time position embedding method called Date2Vec (D2V) and design a novel Fusion Block to enhance the model's utilization of time position embeddings. Thus, we propose the D2Vformer model, as detailed below:
\begin{equation}
\hat{\pmb{Y}} = \text{Fusion} (f(\pmb{X}),\text{D2V}(f(\pmb{X}),\pmb{D}_{x}),\text{D2V}(f(\pmb{X}),\pmb{D}_{y})).
\end{equation}

 The regular time series prediction tasks, depicted on the left side of Figure \ref{preliminary}, involve an input sequence of fixed length 'Seq' and a predicted sequence of fixed length 'Pred', both being adjacent in timestamps. However, in real-world scenarios, the model's predicted sequence length during inference may dynamically change, and the input and predicted sequences may not be adjacent, as shown on the right side of Figure \ref{preliminary}, where the predicted sequence length '$\text{Pred}^*$' varies dynamically, with an interval of length 'Gap' separating the input and predicted sequences. We refer to these tasks as flexible prediction. Furthermore, to prevent confusion, we refer to sequences collected from sensors as 'feature sequences' and sequences representing time information as "timestamps" in the following text.

\section{Methodology}

In general, a time series comprises feature sequences aligned with timestamps. Current methods often focus solely on analyzing the feature sequences, overlooking the valuable information contained in each timestamp. To address this limitation, we introduce a novel approach called Date2Vec(D2V) for time position embedding. This method encodes the position information for each time step and incorporates relationships between different positions implied by the feature sequence. Additionally, we propose a new fusion architecture for integrating position information, enhancing the utilization of time position embedding and feature sequence, thus improving prediction accuracy. 
We refer to this proposed model as D2Vformer, which allows for flexible prediction.

\begin{figure}[t]
    \centering
    \includegraphics[width=1\textwidth]{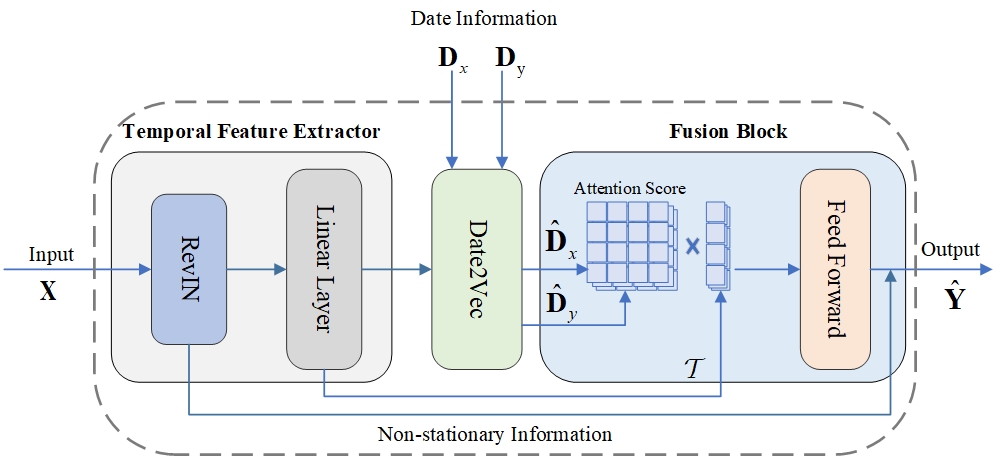}
    \caption{D2Vformer Framework}
    \label{ourmodel}
\end{figure}

\subsection{Overall architecture of D2Vformer}

Figure \ref{ourmodel}  illustrates the overall architecture of our D2Vformer model, which comprises a Temporal Feature Extraction (TFE) module, a Date2Vec (D2V) module, and a Fusion Block. Here, we provide a brief overview of the overall architecture. Detailed explanations of the TFE module, the D2V module, and the Fusion Block can be found in the following subsections.

Unlike most existing time series prediction methods, the D2Vformer model's input includes not only the date matrix $\pmb{D}_{x}$ of the input time series but also the date matrix $\pmb{D}_{y}$ corresponding to the predicted sequence and the feature sequence $\mathbf{X}$. Since the date matrix can be known beforehand, this approach does not lead to data leakage. D2Vformer utilizes the TFE module to capture the features from the feature sequence $\mathbf{X}$ and input them, along with the date matrices $\pmb{D}_{x}$ and $\pmb{D}_{y}$, into the D2V module to generate the time position embeddings. This process can be summarized as follows:
\begin{eqnarray}
    \boldsymbol{\mathcal{T}} &=& \text{TFE}(\pmb{X}),\\
    \hat{\pmb{D}}_{x},\hat{\pmb{D}}_{y} &=& \text{D2V}(\boldsymbol{\mathcal{T}},\pmb{D}_{x},\pmb{D}_{y}),
\end{eqnarray}
where $\boldsymbol{\mathcal{T}} \in \mathbb{R}^{L \times H}$ denotes the features of the input sequence with dimension $H$. $\hat{\pmb{D}}_{x} \in \mathbb{R}^{L \times H \times (k+1)}$ denotes the time position embedding of the input sequence, and $\hat{\pmb{D}}_{y} \in \mathbb{R}^{O \times H \times (k+1)}$ denotes the corresponding time position embedding of the predicted sequence. $(k+1)$ denotes the dimension of the time position embedding, comprising of one linear component and $k$ frequency components. 
The time position embeddings generated by D2V not only contain basic time position information but also encapsulate the relationships between time steps implied by the feature sequence.

In order to fully utilize the time position embeddings and assist the model in achieving flexible prediction, D2Vformer introduces the Fusion Block. Initially, this block models the mapping relationship between the predicted and input sequence based on the time position embeddings. Then, it generates the predicted sequence according to this mapping. Fusion Block is defined as:
\begin{eqnarray}
    \hat{\pmb{Y}} = \text{Fusion}(\boldsymbol{\mathcal{T}},\hat{\pmb{D}}_{x},\hat{\pmb{D}}_{y}),
\end{eqnarray}
where $\hat{\pmb{Y}}$ represents the output of D2Vformer. Unlike the conventional output layer in other models, the Fusion Block in D2Vformer does not predetermine the length of the output sequence, thus enabling flexible prediction. The detials of Fusion Block are provided in Section \ref{Fusion_Block}.

\subsection{Temporal Feature Extraction Module} \label{TFE}

The Temporal Feature Extraction Module (TFE) is designed to extract the features from the input sequences, and the output features will be utilized in D2V. To enhance the extraction of features from the input sequence, TFE incorporates RevIN\cite{revin} as a preprocessing step to eliminate non-stationary information.
By employing RevIN, the TFE module effectively stabilizes the input sequence, thus reducing the complexity of extracting features. 
Inspired by DLinear\cite{dlinear}, TFE also utilizes a linear layer to extract temporal features $\boldsymbol{\mathcal{T}}$, ensuring the simplicity and efficiency of this module.
Specifically, TFE can be represented as:
\begin{eqnarray}
    \boldsymbol{\mathcal{T}} = \text{RevIN}(\pmb{X})\pmb{W}  + \pmb{B},
\end{eqnarray}
where $\pmb{W} \in \mathbb{R}^{D \times H}$ is the weight matrix, $\pmb{B} \in \mathbb{R}^{L \times H}$ is the bias matrix. 
As TFE utilizes RevIN to eliminate non-stationary information from the input time series, it is necessary to restore the non-stationary information\cite{revin} in its output after the Fusion Block completes the prediction.

\subsection{Date2Vec} \label{Date2Vec_Tensor}

\begin{wrapfigure}{r}{0.36\textwidth}
    \centering
    \includegraphics[scale=0.36]{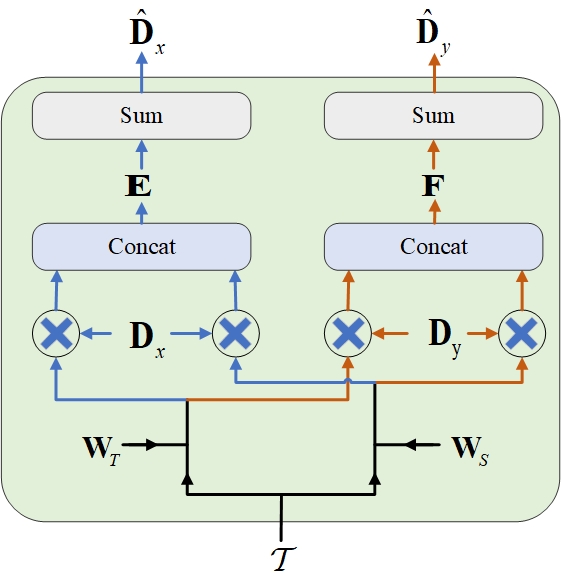}
    \caption{Date2Vec, the blue line represents the process of generating embeddings for the input sequence, while the orange line represents the process of generating embeddings for the predicted sequence.}
    \label{D2V}
\end{wrapfigure}

We introduce a novel method for time position embedding called Date2Vec (D2V). The architecture of D2V is illustrated in Figure \ref{D2V}. This method not only enables the learning of linear patterns in time positions but also facilitates the learning of complex periodic patterns in feature sequences.
To achieve this, D2V incorporates the date matrices $\pmb{D}_{x} \in \mathbb{R}^{L \times M}$ and $\pmb{D}_{y} \in \mathbb{R}^{O \times M}$, along with the temporal feature $\boldsymbol{\mathcal{T}} \in \mathbb{R}^{L \times H}$ as inputs, where $\pmb{D}_{x}$ and $\pmb{D}_{y}$ contain timestamp information for the input and predicted sequences respectively, and $\boldsymbol{\mathcal{T}}$ is extracted from the feature sequences via the TFE module.
Specifically, it can be formulated as:
\begin{eqnarray}
   \pmb v_{T} &=& \pmb{w}_{T} \boldsymbol{\mathcal{T}} + \pmb{b}_{T},\quad
    \pmb \Omega_{S} = \pmb{W}_{S} \boldsymbol{\mathcal{T}} + \pmb{B}_{S},\label{complex components} \\
     \pmb{E} &=& \text{Concat}( \pmb v_{T} \otimes  \pmb{D}_{x}+\pmb{b}_1, \sin(\pmb \Omega_{S} \otimes  \pmb{D}_{x})+\pmb{B}_2 ),\label{time position E} \\ 
      \pmb{F} &=&  \text{Concat}( \pmb v_{T} \otimes  \pmb{D}_{y}+\pmb{b}_3, \sin(\pmb \Omega_{S} \otimes  \pmb{D}_{y})+\pmb{B}_4 )\label{time position F}, 
\end{eqnarray}
where $\pmb{w}_{T} \in \mathbb{R}^{1 \times L}$, $\pmb{b}_{T} \in \mathbb{R}^{H}$, $\pmb{W}_{S} \in \mathbb{R}^{k \times L} $, $\pmb{B}_{S} \in \mathbb{R}^{k \times H} $, $\pmb{b}_{1}$, $\pmb{b}_{3} \in \mathbb{R}^{H}$, $\pmb{B}_{2}$, $\pmb{B}_{4} \in \mathbb{R}^{k \times H}$ are trainable parameters, $ \pmb v_{T} \in \mathbb{R}^{H}$ is the linear component and $\pmb \Omega_{S} \in \mathbb{R}^{k \times H}$ is the frequency component that includes $k$ frequencies. $\otimes$ denotes the Kronecker product, and the concatenation operation is represented as $\text{Concat}(\cdot,\cdot)$. $\pmb{E} \in \mathbb{R}^{(k+1) \times H \times L \times M} $ represents the time position embedding of the input sequence, while $\pmb{F} \in \mathbb{R}^{(k+1) \times H \times O \times M} $ denotes the time position embedding of the prediction. To mitigate the space complexity of D2V, we aggregate the $M$ elements within the time position embedding:
\begin{eqnarray}
    \hat{\pmb{D}}_{x} = \sum_l \pmb{E}_{i,j,k,l},\quad
    \hat{\pmb{D}}_{y} = \sum_l \pmb{F}_{i,j,k,l},
    \label{summation}
\end{eqnarray}
where $\hat{\pmb{D}}_{x} \in \mathbb{R}^{(k+1) \times H \times L}$ is the final time position embedding of the input time series and $\hat{\pmb{D}}_{y} \in \mathbb{R}^{(k+1) \times H \times O}$ is the final time position embedding of the prediction. 

Compared to other methods like T2V\cite{time2vec}, D2V extracts time-step positional relationships by leveraging feature sequence, which not only aligns better with real-world scenarios but also enables the incorporation of intricate patterns within the generated time position embeddings.

\subsection{Fusion Block} \label{Fusion_Block}

Existing models often treat time position embeddings, whether generated by T2V\cite{t2v_position_1,t2v_position_2,t2v_position_3,t2v_position_4} or vanilla Transformer\cite{transformer,fedformer,autoformer}, as mere additional information, integrating them with feature sequences through simple addition or concatenation operations as input to the prediction model.
However, these methods fail to fully utilize the time position embeddings provided by the D2V module. To address this limitation, we propose the Fusion Block, which employs an attention mechanism to learn the similarity between the time position embeddings of the input and predicted sequences, leveraging this similarity to make predictions. The formulation of Fusion Block is as follows:
\begin{eqnarray}
    \pmb{A}_{i,j,l} &= &\sum_k\left(\tilde{\pmb{D}}_{x}\right)_{i,j,k} \left(\tilde{\pmb{D}}_{y}\right)_{i,k,l},\\
    \widetilde{\pmb{Y}}_{i,j} &=&\sum_l \pmb{A}_{i,j,l} \left(\boldsymbol{\mathcal{T}}^{\top}\right)_{i,l} , \\
    \hat{\pmb{Y}} &=& \text{FeedForward}(\widetilde{\pmb{Y}}),
\end{eqnarray}
where $\tilde{\pmb{D}}_{x} \in \mathbb{R}^{H \times L \times (k+1)}$ and $\tilde{\pmb{D}}_{y} \in \mathbb{R}^{H \times (k+1) \times O}$ are the results of permuting the dimensions of $\hat{\pmb{D}}_{x}$ and $\hat{\pmb{D}}_{y}$, respectively, such that $\left(\tilde{\pmb{D}}_{x}\right)_{i,j,k}=\left(\hat{\pmb{D}}_{x}\right)_{j,k,i}$ and $\left(\tilde{\pmb{D}}_{y}\right)_{i,k,l}=\left(\hat{\pmb{D}}_{y}\right)_{k,i,l}$.
The tensor $\pmb{A} \in \mathbb{R}^{H \times O \times L }$ represents the attention scores, where each element denotes the similarity between the input and predicted time steps. 
Then, we multiply the attention scores $\pmb{A}$ by the transposed temporal feature $\boldsymbol{\mathcal{T}^\top}$ to obtain $\widetilde{\pmb{Y}} \in \mathbb{R}^{H \times O}$. 
Finally, we apply $\text{FeedForward}(\cdot)$ to $\widetilde{\pmb{Y}}$ to get the predicted sequence $\hat{\pmb{Y}} \in \mathbb{R}^{D \times O}$, where $\text{FeedForward}(\cdot)$ consists of two fully connected layers to ensure the feature dimension of the predicted sequence meets the requirements.

Existing time series prediction models often require retraining and redeployment to achieve flexible prediction.
In contrast, D2Vformer employs the Kronecker product within its D2V module to generate time position embeddings for both input and output sequences. Then, in the Fusion Block, it computes their similarity using batch matrix multiplication and subsequently multiplies the result with the transposed temporal features through batch matrix multiplication. Finally, the predicted output is obtained via $\text{FeedForward}(\cdot)$. It is worth noting that 
in D2Vformer, the dimensions of all trainable parameters are unaffected by the prediction length $O$. Moreover, none of these operations necessitate predefining the prediction length. Therefore, a key contribution of D2Vformer is its ability to achieve flexible prediction.

\section{Experiments}
To assess the performance of D2Vformer, we conduct regular time series prediction experiments and flexible prediction experiments on six real-world datasets. We compare the experimental results obtained from D2Vformer with several state-of-the-art methods. Additionally, we evaluate the effectiveness of D2V and Fusion Block through dedicated experiments, comparing it against several existing relevant approaches.

\subsection{Settings}
\textbf{Datesets:} The experiments cover datasets from diverse domains such as energy, exchange rates, and healthcare. Specifically, for time series forecasting, we utilize ETT\cite{informer}, Exchange\cite{lstnet} and ILI\cite{autoformer} datasets, as commonly used in previous studies\cite{autoformer,fedformer,patchtst}. The dataset statistics are provided in Appendix \ref{date_Statistics}.
All datasets are divided into training, validation, and test sets in a ratio of 6:2:2. Train/val/test sets are zero-mean normalized with the mean and std of training set,
 following\cite{autoformer,dlinear,dateformer}.

\textbf{Baselines:} We compare D2Vformer and D2V with existing representative models and methods to assess their effectiveness. For time series forecasting, we compare D2Vformer against PatchTST\cite{patchtst}, DLinear\cite{dlinear}, Fedformer\cite{fedformer}, and Autoformer\cite{autoformer}. 

\textbf{Implementation Details:} We use Mean Squared Error (MSE) as the loss function and report Mean Absolute Error (MAE) and MSE as the evaluation metrics. For additional implementation details, please refer to Appendix \ref{experimental_details}.

\begin{small}
\begin{table}
\newcommand{\tabincell}[2]{\begin{tabular}{@{}#1@{}}#2\end{tabular}}
\centering
\footnotesize
\caption{
The results of regular time series forecasting. 
}
\label{multi_ts}
\resizebox{0.85\textwidth}{!}{
\begin{tabular}{p{10mm}p{8mm}p{8mm}p{8mm}p{8mm}p{8mm}p{8mm}p{8mm}p{8mm}p{8mm}p{8mm}p{8mm}} 
\toprule
\multirow{2}{*}{Model}       & \multirow{2}{*}{\tabincell{c}{Pred.\\Length}}  & \multicolumn{2}{c}{D2Vformer}         & \multicolumn{2}{c}{PatchTST}      & \multicolumn{2}{c}{DLinear}           & \multicolumn{2}{c}{Fedformer}       & \multicolumn{2}{c}{Autoformer}   \\ 
\cmidrule(lr){3-4} \cmidrule(lr){5-6} \cmidrule(lr){7-8} \cmidrule(lr){9-10} \cmidrule(lr){11-12}
                               & & MAE             & MSE             & MAE             & MSE             & MAE             & MSE             & MAE             & MSE             & MAE             & MSE             \\    
                               \midrule

\multirow{3}{*}{ETTh1}   & 48     & \textbf{0.3938} 	 & \textbf{0.3550}          & \underline{0.3954}       & \underline{0.3570}         & 0.4036    & 0.3637    & 0.4537 & 0.4102 & 0.4675       & 0.4240         \\
                             & 96     & \textbf{0.4331 }         & \underline{0.4135}    & \underline{0.4344}          & \textbf{0.4094}          & 0.4391    & 0.4159    & 0.4935    & 0.4640 & 0.4944         & 0.4831  \\
                             & 336     & \underline{0.5081} 	& \underline{0.5264}    & 0.5133         & \textbf{0.5084}        & \textbf{0.5033 }   & \textbf{0.5084}   & 0.5573          & 0.5568 & 0.5653    & 0.5757    \\
\midrule
\multirow{3}{*}{ETTh2}   & 48     & \textbf{0.2430}          & \underline{0.1298}          & 0.2443         & 0.1312   & \underline{0.2435 }  & \textbf{0.1293}    & 0.2718          & 0.1562          & 0.2881          & 0.1772         \\
                             & 96     & \underline{0.2702}          & 0.1581          & 0.2703    & \underline{0.1580}   & \textbf{0.2700}        & \textbf{0.1569}        & 0.2986        & 0.1866      & 0.3066          & 0.1942        \\
                             & 336     & \underline{0.3136}        & \underline{0.2043}         & 0.3198    & 0.2097   & \textbf{0.3124}  & \textbf{0.1970}   & 0.3332        & 0.2244      & 0.3315       & 0.2221          \\
\midrule
\multirow{3}{*}{Exchange} & 48     & \underline{0.1112}   & \underline{0.0266}      & 0.1177       & 0.0283      & \textbf{0.1109}      & \textbf{0.0259}         & 0.1832        & 0.0623         & 0.1762       & 0.0578  \\
                             & 96     & \textbf{0.1590}  & \textbf{0.0509}          & 0.1624        & 0.0540        & \underline{0.1602}         & \underline{0.0521}        & 0.2177          & 0.0886         & 0.2161 & 0.0872\\
                             & 336     & \underline{0.3235}  & \underline{0.1965}         & 0.3308        & 0.2087        & \textbf{0.3117}         & \textbf{0.1648}        & 0.3822           & 0.2579         & 0.3794          & 0.2531        \\
\midrule
\multirow{3}{*}{ILI}     & 12     & \textbf{0.5566} & \textbf{0.8991} & \underline{0.6358} & \underline{0.9031} & 0.7162     & 1.0944     & 0.6504  & 0.9282   & 1.1173  & 1.9131  \\
                             & 24     & \textbf{0.6345}   & \textbf{1.0152}
                             & \underline{0.7149}       & 1.3065  & 0.8433    & 1.4334   & 0.7531  & \underline{1.1744}   & 1.0813   & 1.8063   \\
                             & 48     & \textbf{0.6712}  & \textbf{1.0625} &\underline{ 0.7504} & \underline{1.1460} & 0.8994        & 1.6191         & 0.8801   & 1.5022   & 1.0793 & 1.8250 \\
\midrule
\multirow{3}{*}{ETTm1}    & 48     & \textbf{0.3472}   & \textbf{0.2974}    & 0.3544  & 0.3013        & \underline{0.3525}& \underline{0.2975} & 0.4855    & 0.4577    & 0.5358   & 0.5925    \\
                             & 96     & \textbf{0.3662} & \underline{0.3290}   & 0.3802 & 0.3400 & \underline{0.3722} & \textbf{0.3277} & 0.5052 & 0.4950    & 0.5534   & 0.6295          \\
                             & 336     & \textbf{0.4410}   & \underline{0.4644}    & 0.4577         & 0.4773        & \underline{0.4548} & \textbf{0.4582} & 0.5772         & 0.6425   & 0.5947   & 0.7004    \\
\midrule
\multirow{3}{*}{ETTm2}     & 48     & \textbf{0.2038}   & \textbf{0.0934}   & 0.2066    & 0.0948    & \underline{0.2065}        & \underline{0.0946}       & 0.2257         & 0.1066         & 0.2315        & 0.1129        \\
                             & 96     & \textbf{0.2232 }        & \textbf{0.1105}        & 0.2262  & 0.1147        & \underline{0.2260}      & \underline{0.1140}        & 0.2400        & 0.1231        & 0.2455          & 0.1269         \\
                             & 336     & \textbf{0.2774}       & \textbf{0.1713}          & 0.2806 & 0.1738    & \underline{0.2805}       & \textbf{0.1713}       & 0.2830       & \underline{0.1736}    & 0.2910             & 0.1792 \\   
                             \bottomrule
\end{tabular}
}
\end{table}
\end{small}

\subsection{Regular Time Series Forecasting } \label{multivariate}

Table \ref{multi_ts} shows the forecasting accuracy of D2Vformer compared to baseline models on six datasets, with an input length of 96 and varying prediction lengths. 
Notably, due to the small size of the ILI dataset, as per previous time series forecasting studies\cite{autoformer,fedformer,patchtst}, we set an input length of 36 for this dataset. The table highlights the best results in \textbf{bold}, and the second-best results are in \underline{underlined}.
Overall, D2Vformer demonstrates superior performance, achieving either the best or second-best results in MAE and MSE across the majority of cases. 
For instance, in the ETTm2 dataset experiments, D2Vformer outperforms the second-best model, DLinear, by $1.24\%$ in MSE and $3.07\%$ in MAE, for a prediction length of 96. Particularly noteworthy are the substantial improvements D2Vformer achieved in the ILI dataset experiments, with an average enhancement of $11.42\%$ in MAE and $10.01\%$ in MSE compared to the second-best model  across three experiments with different prediction lengths in this dataset.

{\small
\begin{table}[tb]
\centering
\footnotesize
\caption{The results of flexible prediction. 
}
\label{skip_exp}
\resizebox{0.95\textwidth}{!}{
\begin{tabular}{p{10mm}p{6mm}p{6mm}p{6mm}p{8mm}p{8mm}p{8mm}p{8mm}p{8mm}p{8mm}p{8mm}p{8mm}p{8mm}p{8mm}}
 
\toprule
\multirow{2}{*}{Model}       & \multicolumn{3}{c}{Configs}  & \multicolumn{2}{c}{D2Vformer}         & \multicolumn{2}{c}{PatchTST}      & \multicolumn{2}{c}{DLinear}   & \multicolumn{2}{c}{Fedformer}        & \multicolumn{2}{c}{Autoformer}   \\ 
\cmidrule(lr){2-4}
\cmidrule(lr){5-6} \cmidrule(lr){7-8} \cmidrule(lr){9-10} \cmidrule(lr){11-12}  \cmidrule(lr){13-14}
                              & Seq & Gap & $\text{Pred}^*$ & MAE             & MSE             & MAE             & MSE             & MAE             & MSE             & MAE             & MSE               & MAE             & MSE           \\    
                               \midrule
\multirow{3}{*}{ILI} 
&36 &4 &2 &\textbf{0.4882}  &\textbf{0.5913} &\underline{0.5081} &\underline{0.6687} &0.5630   &0.7667&0.6863 &	1.0934    &0.8375  &1.4528\\

&36 &4 &4 &\textbf{0.5704}  &\textbf{0.7939}&\underline{0.5849} &\underline{0.8155} &0.6634   &0.9840&0.7633 &	1.3078    &1.0013  &1.8697 \\

 &36 &4 &8 &0.7933  &1.4370&\textbf{0.6690}
&\textbf{1.1007} &0.7829   &\underline{1.2243} &\underline{0.7741} &	1.3439 
  &0.9219  &1.6797 \\

\midrule
\multirow{3}{*}{ILI} 
&36 &6 &2 &\textbf{0.6064}  &\textbf{0.8539} &\underline{0.6375} &\underline{0.9280} &0.8340   &1.3704 &0.8212 &	1.4519   &1.0613  &2.1226  \\

&36 &6 &4 &\textbf{0.6649} &\textbf{1.0015} &\underline{0.6938} &\underline{1.1164} &0.8231  &1.3354 &0.8343 	&1.5232  &0.9732  &1.8404 \\

&36 &6 &8 &0.8810 &1.6633 &\textbf{0.7051} &\textbf{1.1699} &0.8694  &\underline{1.4156} &\underline{0.8593} &1.5569  &0.9795  &1.9084 \\

\midrule
\multirow{3}{*}{ILI} 
&36 &8 &2 &\textbf{0.7238}  &\textbf{1.1830} &\underline{0.7354} &\underline{1.2210} &0.8755   &1.4756 &0.8747 &	1.6337   &0.9963  &1.8991 \\

&36 &8 &4 &\textbf{0.7277}  &\textbf{1.1608} &\underline{0.7472} &\underline{1.3183} &0.8862   &1.4640 &0.8342 &	1.5068   &0.9570  &1.7790 \\

&36 &8 &8 &\underline{0.8331} &\underline{1.5262} &\textbf{0.7572} &\textbf{1.2718} &0.9111   &1.5630 &0.9199 &1.7066  &1.1434  &2.4403 \\

\midrule

\multirow{3}{*}{Exchange} 
&96  &6&4 &\textbf{0.0714}   &\textbf{0.0113} &\underline{0.0717}&0.0117 &0.0718 
   &\underline{0.0114} &0.1312 &	0.0336  &0.1466   &0.0412 \\

&96&6 &8 & \textbf{0.0785}&\textbf{0.0135}&\underline{0.0788}&{0.0140}&0.0793 &\underline{0.0136}&0.1329 &	0.0342  &0.1613 &0.0487 
\\

&96 &6 &10& \textbf{0.0818}&\textbf{0.0147}&0.0822&\underline{0.0151}&\underline{0.0820}&\textbf{0.0147}&0.1371 &	0.0364 & 0.1488 &0.0418 
\\
\midrule
\multirow{3}{*}{Exchange} 
&96  &8&4 &\textbf{0.0785}   &\textbf{0.0136} &0.0790 &0.0140 
 &\underline{0.0788} 
   &\underline{0.0137} &0.1357 &0.0359 &0.1615   &0.0490\\

&96&8 &8 & \textbf{0.0854}&\textbf{0.0157}&0.0860 
&0.0162 &\underline{0.0857} &\underline{0.0158}&0.1393 &0.0375 &0.1526  &0.0438  
\\

&96 &8 &10& \textbf{0.0886}&\textbf{0.0165}&0.0893 &0.0172&\underline{0.0888}&\underline{0.0168}&0.1401 &0.0377 &0.1502 &0.0430
\\

\midrule
\multirow{3}{*}{Exchange} 
&96  &10&4 &\textbf{0.0858}&\textbf{0.0157}&0.0864&0.0162&\underline{0.0862}&\underline{0.0158}&0.1384 &0.0368 &0.1679 &0.0524\\

&96&10 &8& \textbf{0.0921}&\underline{0.0179}&0.0928&0.0183&\underline{0.0923}&\textbf{0.0178}&0.1426 &0.0389 &0.1536&0.0450
\\

&96 &10 &10&\textbf{0.0949}&\textbf{0.0186} &0.0959&0.0194&\underline{0.0953}&\underline{0.0188}&0.1450 &0.0396  &0.1581 &0.0459 \\

                             \bottomrule

\end{tabular}
}
\end{table}
}

\subsection{Flexible Prediction } \label{skip_prediction}
Thanks to the D2V module and the Fusion Block, our D2Vformer can predict a time series with arbitrary positions and lengths by simply inputting the corresponding date vectors. Therefore, in flexible prediction scenarios where the positions or lengths of predictions vary, D2Vformer significantly reduces training costs, as it only requires a single training session. 

Next, we evaluate the performance of D2Vformer in flexible prediction. 
As illustrated in the right figure of Figure \ref{preliminary}, in this experiment, the input sequence and the predicted sequence are not adjacent. There is a specific length of interval between them. Hence, we have three key parameters for this experiment: (1) 'Seq' representing the length of the input sequence, (2) '$\text{Pred}^*$' representing the length of the predicted sequence, and (3) 'Gap' indicating the length of the interval, which separates the input and predicted sequences. 
During training, D2Vformer learns from a regular time series prediction task where the input and prediction sequences are adjacent,  and the prediction length is set to 'Gap.' During inference, there's an interval of length 'Gap' between the input and prediction sequences. In this scenario, we require D2Vformer to directly generate predictions of varying lengths.
Since other baseline models cannot directly perform interval-based prediction, we extend the length of their predicted sequences to 'Gap'+'$\text{Pred}^*$' and extract the last '$\text{Pred}^*$' length of data as the prediction results for comparison with D2Vformer.

Table \ref{skip_exp} presents the experimental results of D2Vformer and other baseline models in flexible prediction scenarios, where the best results are highlighted in \textbf{bold}, and the second-best results are \underline{underlined}. 
From the table, it is evident that in short-term prediction tasks, D2Vformer outperforms other baseline models in terms of MAE and MSE in the majority of cases.
For instance, in the case of the ILI dataset, D2Vformer demonstrates a reduction in mean squared error (MSE) compared to the suboptimal model, with respective improvements of $10.29\%$ and $11.95\%$ observed for experiments involving a '$\text{Pred}^*$' of 4 and 'Gap's of 6 and 8. Importantly, in experiments with the same 'Gap', D2Vformer undergoes a single training session to generate predictions for varying '$\text{Pred}^*$'s, resulting in significant time savings in training. 
For medium- to long-term prediction tasks, the corresponding flexible prediction results can be found in the Appendix \ref{flexible_prediction}.

\begin{table}[tb]
\centering
\footnotesize
\newcommand{\tabincell}[2]{\begin{tabular}{@{}#1@{}}#2\end{tabular}}
\caption{Performance of time position embedding methods. 
}
\label{D2V_position}
\resizebox{0.55\textwidth}{!}{
\begin{tabular}
{p{13mm}p{9mm}p{8mm}p{8mm}p{8mm}p{8mm}p{8mm}p{8mm}} 
\toprule
\multirow{2}{*}{Model}       & \multirow{2}{*}{\tabincell{c}{Pred.\\Length}}   & \multicolumn{2}{c}{D2V}         & \multicolumn{2}{c}{T2V}      & \multicolumn{2}{c}{Primal Position}             \\ 
\cmidrule(lr){3-4} \cmidrule(lr){5-6} \cmidrule(lr){7-8} 
                              &  & MAE             & MSE             & MAE             & MSE             & MAE             & MSE                  \\    
                               \midrule

\multirow{2}{*}{Exchange}   & 96     & \textbf{0.2949} 	 & \textbf{0.1404}          & 0.3183      & 0.1657         & \underline{0.3033}    & \underline{0.1493}            \\
& 336     & \textbf{0.4919} 	 & \textbf{0.3826}          & \underline{0.6831}      &\underline{0.6748}         & {0.9205}    & {1.5402}            \\
\midrule
\multirow{2}{*}{ETTh1}  & 96     & \underline{0.4598} 	 & \underline{0.4387}          & 0.5349     & 0.5220          & \textbf{0.4573}    & \textbf{0.4292}            \\
& 336     & \textbf{0.6128}	 & \textbf{0.6742}         & 0.6210    & 0.7323        & \underline{0.6132}  & \underline{0.7101}            \\
\midrule
\multirow{2}{*}{ETTm1}  & 96     & \textbf{0.3965} 	 & \textbf{0.3529}          & 0.4756    & 0.4407      & \underline{0.4160}    & \underline{0.3706 }            \\
& 336     & \textbf{0.5076} 	 & \textbf{0.5196}          & 0.5267    & 0.5608      & \underline{0.5158}    & \underline{0.5569}            \\

                             \bottomrule

\end{tabular}
}
\end{table}
\begin{table}[tb]
\centering
\footnotesize
\newcommand{\tabincell}[2]{\begin{tabular}{@{}#1@{}}#2\end{tabular}}
\caption{Performance of different output modules.
}
\label{Fusion_Block_test}
\resizebox{0.55\textwidth}{!}{
\begin{tabular}
{p{14mm}p{10mm}p{7mm}p{7mm}p{7mm}p{7mm}p{7mm}p{7mm}@{\hspace{2em}}} 
\toprule
\multirow{2}{*}{Model}       & \multirow{2}{*}{\tabincell{c}{Pred.\\Length}}   & \multicolumn{2}{c}{Fusion Block}         & \multicolumn{2}{c}{Attention}      & \multicolumn{2}{c}{Linear}             \\ 
\cmidrule(lr){3-4} \cmidrule(lr){5-6} \cmidrule(lr){7-8} 
                              &  & MAE             & MSE             & MAE             & MSE             & MAE             & MSE                  \\    
                               \midrule

\multirow{3}{*}{Exchange} 
 & 48     & \textbf{0.1135} 	 & \textbf{0.0271}         & \underline{0.1138}      & \underline{0.0275}         & {0.1139}    & \underline{0.0275}            \\
& 96     & \textbf{0.1590} 	 & \textbf{0.0510}         & 0.1635      & 0.0536         & \underline{0.1630}    & \underline{0.0535}    \\

 & 336     & \textbf{0.3262}	 & \textbf{0.2000}       & 0.3286     &    0.2067    & \underline{0.3282}    & \underline{0.2012}           \\
\midrule

\multirow{3}{*}{ETTh1} 
&48     & \textbf{0.3940} 	 & \underline{0.3554}         & \underline{0.3944}      & 0.3567      & \underline{0.3944}   & \textbf{0.3550}           \\
 & 96     & \textbf{0.4331} 	 & \underline{0.4136}         & \underline{0.4340}      & 0.4144        & 0.4347   & \textbf{0.4117}           \\
 & 336     & \textbf{0.5099} 	 & \textbf{0.5286}       & \underline{0.5101}     & 0.5413        & 0.5111    & \underline{0.5306}       \\

\midrule
\multirow{3}{*}{ETTm1}  
 & 48     & \textbf{0.3471} 	 & {0.2980}         & {0.3502}      & \underline{0.2979}       & \underline{0.3477}   & \textbf{0.2968}         \\     
 & 96     & \textbf{0.3662} 	 & \textbf{0.3293}         & \underline{0.3669}      & \underline{0.3297}       & 0.3679   & 0.3335         \\
 & 336     & \textbf{0.4407} 	 & \underline{0.4637}         & \underline{0.4422}      & {0.4641}       & 0.4431   & \textbf{0.4609}         \\ 
                             \bottomrule
\end{tabular}
}
\end{table}

\subsection{The Effectiveness of D2V \& Fusion Block} 
We further investigate the effectiveness of the D2V and Fusion Block. 
To validate the effectiveness of D2V, we compared it with two alternative algorithms, T2V \cite{t2v_position_1} and the position embedding method used in vanilla Transformer \cite{transformer}, as shown in Table \ref{D2V_position}.
It is observed that the performance of D2V is the best in the majority of cases.
To assess the impact of different output modules, we replace the Fusion Block with other commonly used output modules. The experimental results are presented in Table \ref{Fusion_Block_test}. From the table, it is evident that the Fusion Block outperforms other modules in most prediction scenarios. 
Furthermore, due to its inherent structure, the Fusion Block can meet the requirements of flexible prediction tasks, providing a significant advantage in output flexibility compared to conventional output modules.

\subsection{Efficiency Analysis}

\begin{wraptable}{r}{0.55\textwidth}
    \footnotesize
    \caption{Evaluation of model complexity in regular and flexible prediction tasks. 'Numb. of Param.' refers to the number of learnable parameters in the model, 'Inference' indicates the average time taken to make predictions for a batch of data, and 'Training' denotes the average time taken to train the model one epoch.}
    \label{computation_efficiency}
    \newcommand{\tabincell}[2]{\begin{tabular}{@{}#1@{}}#2\end{tabular}}
    \begin{subtable}{0.47\textwidth}
        \centering
        \resizebox{1\textwidth}{!}{
        \begin{tabular}
            {ccccccc}  
            \toprule
            \multirow{2}{*}{Model}    & \multicolumn{2}{c}{Configs}     & \multirow{2}{*}{\tabincell{c}{Numb. of\\ Param.}}        & \multirow{2}{*}{\tabincell{c}{Inference \\(ms/batch)}}      & \multirow{2}{*}{\tabincell{c}{Training \\(s/epoch)}}            \\ 
            \cmidrule{2-3}
            & Seq  & Pred \\
            \midrule
            D2Vformer
            & 96     & 96 	 & 146038          & 7.4910           & 2.4687     \\
            PatchTST  
            & 96     & 96 	 & 6310336          & 5.9510          & 6.1480      \\
            DLinear 
            & 96     & 96 	 & 19012         & 3.6151         & 0.6950    \\
            Autoformer  
            & 96     & 96 	 & 7330252        & 78.9015          & 13.3086      \\
            Fedformer 
            & 96     & 96 	 & 22589454        & 79.9172        & 29.6838  \\
            \bottomrule
        \end{tabular}}
        \caption{On regular prediction}
        \label{computation_mul}
    \end{subtable}\\
    \begin{subtable}{0.47\textwidth}
        \centering
        \resizebox{1.1\textwidth}{!}{
            \begin{tabular}
                {ccccccc} 
                \toprule
                \multirow{2}{*}{Model}   & \multicolumn{3}{c}{Configs}     & \multirow{2}{*}{\tabincell{c}{Numb. of\\ Param.}}        & \multirow{2}{*}{\tabincell{c}{Inference \\(ms/batch)}}      & \multirow{2}{*}{\tabincell{c}{Training \\(s/epoch)}}        \\ 
                \cmidrule{2-4}
                &Gap & Seq  & $\text{Pred}^*$\\
                \midrule
                \multirow{3}{*}{D2Vformer}   &\multirow{3}{*}{6} & \multirow{3}{*}{96}     & 4 & 146038       & 5.5140 & 1.5818               \\
                & & &8 & -       & 6.8280 & -              \\
                & &  &10 & -       & 7.0502 & -               \\
                \midrule
                \multirow{3}{*}{PatchTST}  &\multirow{3}{*}{6} & \multirow{3}{*}{96}     & 4  & 4425288      & 3.8227  & 6.0567              \\
                 & &   &8  & 4510972      & 3.9979  & 6.0660               \\
                  & &    &10  & 4553814      & 4.1038 &6.0786                \\
                  \midrule
                \multirow{3}{*}{DLinear}  &\multirow{3}{*}{6} & \multirow{3}{*}{96}     & 4  & 1940      & 3.0557 & 0.6509          \\
                 & &   & 8  & 2716      & 3.1221 & 0.6557          \\
                  & &   & 10  & 3104      & 3.1863 & 0.6631          \\
                  \midrule
                \multirow{3}{*}{Autoformer}  &\multirow{3}{*}{6} & \multirow{3}{*}{96}    & 4  & 7330252      & 66.9880 & 7.2816           \\
                 & &  & 8  & 7330252      & 68.4921 & 7.5508          \\
                 & &    & 10  & 7330252     & 69.2782 & 7.9322          \\
                 \midrule
                \multirow{3}{*}{Fedformer}  &\multirow{3}{*}{6} & \multirow{3}{*}{96}     & 4  & 16520454      &58.3564  & 18.5534           \\
                 & &    & 8  & 16867254      &59.3064 & 18.8975           \\
                  & &    & 10  & 17040654      &60.4832  & 19.1533           \\
                \bottomrule
            \end{tabular}}
        \caption{On flexible prediction}
        \label{computation_skip}
    \end{subtable}
\end{wraptable}

In Section \ref{skip_prediction}, we mentioned that one major advantage of D2Vformer is its significant reduction in training time, particularly in variable-length prediction tasks where D2Vformer does not require multiple training deployments. To validate this advantage of D2Vformer, in this section, we comprehensively analyze the computational efficiency of D2Vformer. All the experimental settings align with the corresponding experiments outlined in Section \ref{multivariate} and Section \ref{skip_prediction}.

As shown in Table \ref{computation_efficiency}, we have evaluated the computational efficiency of D2Vformer and baseline models across various prediction scenarios using three key metrics, including (1) Parameters, indicating the number of learnable parameters; (2) Inference, representing the average inference time for a batch of data; (3) Training, denoting the average time required for training one epoch. It is worth noting that all experimental results presented are the average results of 50 trials.

The number of parameters in D2Vformer is independent of the length of the output time series, giving it a significant advantage over baseline models in terms of the number of learnable parameters. For the prediction tasks with a prediction length of 96, as shown in Table \ref{computation_efficiency}(a), D2Vformer has 146,038 parameters, second only to the simplest model, DLinear.
Additionally, in Table \ref{computation_efficiency}(b), for different settings of '$\text{Pred}^*$,' D2Vformer's average training time for an epoch is also second only to the simplest model, DLinear. Furthermore, in scenarios where the prediction length dynamically changes during the inference stage, D2Vformer does not require multiple training deployments like other comparison models. Therefore, in such cases, D2Vformer has a significant advantage in training time compared to baseline models.
Furthermore, in Appendix \ref{Effectiveness1}, we conduct supplementary experiments to evaluate the effectiveness of D2Vformer and other methods in various flexible prediction scenarios.

\section{Conclusion}
Time position is crucial for distinguishing time steps and representing the sequence order. However, preliminary experiments indicate that current methods do not fully leverage this information in time series forecasting. To address this, this paper introduces a novel time position embedding method called D2V and a new time series prediction model named D2Vformer.
D2V efficiently generates time position embeddings by capturing both linear and periodic positional relationships among time steps based on timestamps and feature sequences. To fully utilize these embeddings, we propose a new Fusion Block, enabling D2Vformer to excel not only in regular time series prediction tasks but also in flexible prediction tasks with intervals and varying prediction lengths. Experimental results on both regular and flexible prediction tasks demonstrate the superiority of D2Vformer.
In flexible prediction tasks, D2Vformer requires only a single training session to predict varying-length sequences during inference. This capability is particularly suitable for dynamically changing prediction scenarios, leading to significant savings in training costs and time. We believe that this work can inspire further research in the field of time position embeddings for time series modeling.

\newpage

\bibliographystyle{unsrt}
\bibliography{arxiv}

\newpage

\appendix

\section{Appendix}\label{appendix_a}

In this section, we provide additional information and results from supplementary experiments that support the main body of the paper. 

\subsection{Date Embedding}\label{date_embedding}

Our method to generate date vectors follows the methodology proposed by Dateformer\cite{dateformer}, which incorporates the following features for a given date at a specific time step: abs\_day, year, day (month day), year day, weekofyear, lunar\_year, lunar\_month, lunar\_day, lunar\_year\_day, dayofyear, dayofmonth, monthofyear, dayofweek, dayoflunaryear, dayoflunarmonth, monthoflunaryear, jieqiofyear, jieqi\_day, dayofjieqi. These features are encoded using the following equations:
\begin{eqnarray}
    abs\_day &=& \frac{\text{days that have passed from December 31, 2000}}{365.25*5} \nonumber \\
    year &=& \frac{\text{this year} - 1998.5}{25} \nonumber \\
    day &=& \frac{\text{days that have passed in this month}}{31}\nonumber \\
    year\_day &=& \frac{\text{days that have passed in this year}}{366} \nonumber \\
    weekofyear&=& \frac{\text{weeks that have passed in this year}}{54} \nonumber \\
    lunar\_year&=& \frac{\text{this lunar year} - 1998.5}{25} \nonumber \\
    lunar\_month&=& \frac{\text{this lunar month}}{12} \nonumber \\
    lunar\_day&=& \frac{\text{days that have passed in this lunar month}}{30} \nonumber \\
    lunar\_year\_day&=& \frac{\text{days that have passed in this lunar year}}{384} \nonumber \\
     dayofyear&=& \frac{\text{days that have passed in this year} - 1}{\text{total days in this year} - 1}-0.5 \nonumber \\
     dayofmonth&=& \frac{\text{days that have passed in this month} - 1}{\text{total days in this month} - 1}-0.5 \nonumber \\
     monthofyear&=& \frac{\text{months that have passed in this year} - 1}{11}-0.5 \nonumber \\
     dayofweek&=& \frac{\text{days that have passed in this week} - 1}{6}-0.5 \nonumber \\
     dayoflunaryear&=& \frac{\text{days that have passed in this lunar year}-1}{\text{total days in this lunar year}-1}-0.5 \nonumber \\
     dayoflunarmonth&=& \frac{\text{days that have passed in this lunar month} - 1}{\text{total days in this lunar month} - 1}-0.5 \nonumber \\
     monthoflunaryear&=& \frac{\text{lunar months that have passed in this lunar year} - 1}{11}-0.5 \nonumber \\
     jieqiofyear&=& \frac{\text{solar terms that have passed in this year} - 1}{23}-0.5 \nonumber \\
     jieqi\_day&=& \frac{\text{days that have passed in this solar term}}{15}\nonumber \\
     dayofjieqi&=& \frac{\text{days that have passed in this solar term} - 1}{\text{total days in this solar term} - 1}-0.5\nonumber
    \end{eqnarray}

\subsection{Date Statistics}\label{date_Statistics}
\begin{small}
\begin{table}[!t]
\centering
\footnotesize
\caption{Dataset Statistics}
\label{Commonly used datasets}
\begin{tabular}{p{25mm}p{18mm}p{18mm}p{20mm}}
\toprule
Datasets & Timesteps  &  Variates & Granularity \\
\midrule
ETTm1\&ETTm2 &69680   &7   &5 min  \\
\midrule
ETTh1\&ETTh2 &17420   &7  &1 hour  \\
\midrule
Exchange &7588   &8  &1 day  \\
 \midrule
ILI  &966   &7  &1 week  \\
\bottomrule
\end{tabular}
\end{table}
\end{small}

We use six popular multivariate datasets from different domains in our experiments. Table \ref{Commonly used datasets} shows an overview of the datasets. In the following, we briefly describe the individual datasets.

\begin{itemize}
    \item[$\bullet$] \textbf{ETT} (Electricity Transformer Temperature)\cite{informer}  encompasses data collected from power transformers in a specific region of China, spanning from July 2016 to July 2018. It includes seven distinct features, such as the load and oil temperature of the power transformers.  ETTh1 and ETTh2  consist of 17,420 samples with a sampling interval of 1 hour, encompassing 7 feature dimensions.  ETTm1 and ETTm2 comprise 69,680 samples with a sampling interval of 5 minutes, encompassing 7 feature dimensions.

    \item[$\bullet$] \textbf{Exchange}\cite{lstnet} covers daily exchange rate data for eight currencies for the period January 1990 to October 2010, and it contains eight feature dimensions. The dataset has a sampling interval of 1 day and contains a total of 7588 samples.

    \item[$\bullet$]  \textbf{ILI} (Influenza-Like Illness)\cite{autoformer} encompasses weekly records of patients diagnosed with influenza illness, as reported by the U.S. Centers for Disease Control and Prevention, spanning from 2002 to 2021. This dataset comprises seven key dimensions of characterization, including features such as the proportion of patients with ILI and the total number of patients. It follows a sampling interval of 1 week and comprises a total of 966 samples.

\end{itemize}

\subsection{Additional Experimental Details}\label{experimental_details}

All experiments are conducted using the PyTorch framework \cite{pytorch} on a single NVIDIA RTX2080TI 11GB GPU. MSE is used as the loss function, Adam \cite{adam} is used as the optimizer, and the mini-batch size is set to 16 for all models. The training is capped at 100 epochs, with early stopping applied after 5 epochs of no improvement.
To minimize the impact of the initial learning rate ($\hat{lr}$) on model performance, we experiment with six different initial learning rates: $\left \{ 10^{-3}, 5\times10^{-4}, 10^{-4}, 5\times10^{-5}, 10^{-5}, 5\times10^{-6} \right \}$. Each model is then evaluated using its optimal learning rate for comparison.
Additionally, the learning rate ($lr$) decay strategy for all models is defined as: $$lr = \hat{lr} \times 0.75^{\left \lfloor \frac{\text{epoch}-1}{2}\right \rfloor}.$$
For each Transformer-based comparison model, we use their common parameter settings: the encoders stack 2 layers, the decoder stacks 1 layer, the hidden dimensions are set to $\left\{ 512, 1024 \right\}$, and the Dropout rate is $0.1$. For D2Vformer, the output dimension $H$ of the Feature Extraction (TFE) module is set to $512$, and the number of frequency components $k$ in the D2V module is set to $63$. Each experiment is repeated five times with different random seeds, and the average of the five prediction results is reported as the model's final performance.

\subsection{Baselines}\label{baselines}

We conduct comparative experiments with four state-of-the-art time series prediction models. Below is a brief introduction to each of these approaches.

\begin{itemize}
    \item [$\bullet$] \textbf{PatchTST}\cite{patchtst} is a Transformer-based model that employs patch slicing method. For the ILI dataset, given its limited number of samples, we set the patch length to 6 with a stride of 2. For the other datasets, we set the patch length to 16 with a stride of 2.

    \item [$\bullet$] \textbf{DLinear}\cite{dlinear} is a prediction model based on linear layers and time series decomposition, where time series decomposition is composed of mean filters. We used the implementation available in DLinear with the recommended hyperparameters.

    \item [$\bullet$] \textbf{Fedformer}\cite{fedformer} is a Transformer-based model that transfers the input time series into the frequency domain, randomly drops frequency components, and calculates attention relationships among frequency components. The number of preserved frequency components is given by $\min(64, \text{Seq}/2)$. We used the implementation available in Fedformer with the recommended hyperparameters.

    \item [$\bullet$] \textbf{Autoformer}\cite{autoformer} is a Transformer-based model that proposes the AutoCorrelation module. This module explores potential periodic patterns by calculating the temporal lag similarity of the time series and retains the top $U$ factors based on their similarity scores. We set $U = 5*\ln_{}{\text{Seq}}$. We used the implementation available in Autoformer with the recommended hyperparameters.
    
\end{itemize}

{\small
\begin{table}[tb]
\centering
\footnotesize
\caption{The results of short-term interval-based prediction. The experiments are conducted for the two datasets using the four baselines. We report the average value and standard deviation of experiments.
}
\label{deviation}
\resizebox{1\textwidth}{!}{
\begin{tabular}{p{10mm}p{2mm}p{2mm}p{4mm}p{10mm}p{10mm}p{10mm}p{10mm}p{10mm}p{10mm}p{10mm}p{10mm}p{10mm}p{10mm}}
 
\toprule
\multirow{2}{*}{Model}       & \multicolumn{3}{c}{Configs}  & \multicolumn{2}{c}{D2Vformer}         & \multicolumn{2}{c}{PatchTST}      & \multicolumn{2}{c}{DLinear}   & \multicolumn{2}{c}{Fedformer}        & \multicolumn{2}{c}{Autoformer}   \\ 
\cmidrule(lr){2-4}
\cmidrule(lr){5-6} \cmidrule(lr){7-8} \cmidrule(lr){9-10} \cmidrule(lr){11-12}  \cmidrule(lr){13-14}
                              & Seq & Gap & $\text{Pred}^*$ & MAE             & MSE             & MAE             & MSE             & MAE             & MSE             & MAE             & MSE               & MAE             & MSE           \\    
                               \midrule

\multirow{6}{*}{ILI} 
&\multirow{6}{*}{36} &\multirow{6}{*}{4} &\multirow{2}{*}{2}&\textbf{0.4882}  &\textbf{0.5913} &\underline{0.5081} &\underline{0.6687} &0.5630   &0.7667&0.6863 &	1.0934    &0.8375  &1.4528\\

& & & & $\pm 0.0123$  &$\pm 0.0168$ &$\pm 0.0154$&$\pm 0.0491$ &$\pm 0.0061 $ &$\pm 0.0262$&$\pm 0.0212 $&	$\pm 0.0676  $ &$\pm 0.0071 $ &$\pm 0.0347$\\

& & &\multirow{2}{*}{4}  &\textbf{0.5704}  &\textbf{0.7939}&\underline{0.5849} &\underline{0.8155} &0.6634   &0.9840&0.7633 &	1.3078    &1.0013  &1.8697 \\

& & & &$\pm 0.0083  $&$\pm 0.0041$ &$\pm 0.0083$&$\pm 0.2013$ &$\pm 0.0174 $ &$\pm 0.0816$&$\pm 0.0117 $&$	\pm 0.0252 $  &$\pm 0.0345$  &$\pm 0.0652$\\

 & & &\multirow{2}{*}{8}  &0.7933  &1.4370&\textbf{0.6690}
&\textbf{1.1007} &0.7829   &\underline{1.2243} &\underline{0.7741} &	1.3439 
  &0.9219  &1.6797 \\
& & & &$\pm 0.0045  $&$\pm 0.0087 $&$\pm 0.0074$&$\pm 0.0180 $&$\pm 0.0112  $&$\pm 0.1285$&$\pm 0.0095 $&$	\pm 0.0364   $&$\pm 0.0061 $ &$\pm 0.0960$\\


\midrule
\multirow{6}{*}{ILI} 
&\multirow{6}{*}{36} &\multirow{6}{*}{6} &\multirow{2}{*}{2} &\textbf{0.6064}  &\textbf{0.8539} &\underline{0.6375} &\underline{0.9280} &0.8340   &1.3704 &0.8212 &	1.4519   &1.0613  &2.1226  \\
& & & &$\pm 0.0100  $&$\pm 0.0245$&$\pm 0.0600$&$\pm 0.0117 $&$\pm 0.0112  $&$\pm 0.1461 $&$\pm 0.0095 $&$	\pm 0.0295   $&$\pm 0.0200  $&$\pm 0.0100$\\

& & &\multirow{2}{*}{4} &\textbf{0.6649} &\textbf{1.0015} &\underline{0.6938} &\underline{1.1164} &0.8231  &1.3354 &0.8343 	&1.5232  &0.9732  &1.8404 \\
& & & &$\pm 0.0024  $&$\pm 0.0359 $&$\pm 0.0084$&$\pm 0.0162 $&$\pm 0.0070  $&$\pm 0.0137$&$\pm 0.0093 $&$	\pm 0.0286   $&$\pm 0.0145  $&$\pm 0.0468$\\

& & &\multirow{2}{*}{8} &0.8810 &1.6633 &\textbf{0.7051} &\textbf{1.1699} &0.8694  &\underline{1.4156} &\underline{0.8593} &1.5569  &0.9795  &1.9084 \\
& & & &$\pm 0.0019  $&$\pm 0.0295 $&$\pm 0.0030$&$\pm 0.0036 $&$\pm 0.0106  $&$\pm 0.0441$&$\pm 0.0188 $&$	\pm 0.0405   $&$\pm 0.0049  $&$\pm 0.0558$\\

\midrule
\multirow{6}{*}{ILI} 
&\multirow{6}{*}{36} &\multirow{6}{*}{8} &\multirow{2}{*}{2} &\textbf{0.7238}  &\textbf{1.1830} &\underline{0.7354} &\underline{1.2210} &0.8755   &1.4756 &0.8747 &	1.6337   &0.9963  &1.8991 \\

& & & &$\pm 0.0042  $&$\pm 0.0605 $&$\pm 0.0144$&$\pm 0.0422 $&$\pm 0.0058  $&$\pm 0.0571$&$\pm 0.0422 $&$	\pm 0.0880   $&$\pm 0.0179  $&$\pm 0.0183$\\

& & &\multirow{2}{*}{4} &\textbf{0.7277}  &\textbf{1.1608} &\underline{0.7472} &\underline{1.3183} &0.8862   &1.4640 &0.8342 &	1.5068   &0.9570  &1.7790 \\

& & & &$\pm 0.0115  $&$\pm 0.0396 $&$\pm 0.0124$&$\pm 0.0054 $&$\pm 0.0134  $&$\pm 0.0544$&$\pm 0.0484 $&$	\pm 0.1836   $&$\pm 0.0094  $&$\pm 0.0398$\\

& & &\multirow{2}{*}{8} &\underline{0.8331} &\underline{1.5262} &\textbf{0.7572} &\textbf{1.2718} &0.9111   &1.5630 &0.9199 &1.7066  &1.1434  &2.4403 \\

& & & &$\pm 0.0069 $&$\pm 0.0098 $&$\pm 0.0055$&$\pm 0.0165 $&$\pm 0.0094  $&$\pm 0.1346$&$\pm 0.0108 $&$	\pm 0.1413   $&$\pm 0.0664  $&$\pm 0.1142$\\

\midrule

\multirow{6}{*}{Exchange} 
&\multirow{6}{*}{96}  &\multirow{6}{*}{6} &\multirow{2}{*}{4} &\textbf{0.0714}   &\textbf{0.0113} &\underline{0.0717}&0.0117 &0.0718 
   &\underline{0.0114} &0.1312 &	0.0336  &0.1466   &0.0412 \\

& & & &$\pm 0.0020  $&$\pm 0.0004 $&$\pm 0.0034$&$\pm 0.0009 $&$\pm 0.0081  $&$\pm 0.0019$&$\pm 0.0072 $&$	\pm 0.0032   $&$\pm 0.0073  $&$\pm 0.0033$\\

& & &\multirow{2}{*}{8} & \textbf{0.0785}&\textbf{0.0135}&\underline{0.0788}&{0.0140}&0.0793 &\underline{0.0136}&0.1329 &	0.0342  &0.1613 &0.0487 
\\
& & & &$\pm 0.0027  $&$\pm 0.0008 $&$\pm 0.0011$&$\pm 0.0005 $&$\pm 0.0069  $&$\pm 0.0017$&$\pm 0.0020 $&$	\pm 0.0011   $&$\pm 0.0033  $&$\pm 0.0009$\\

& & &\multirow{2}{*}{10}& \textbf{0.0818}&\textbf{0.0147}&0.0822&\underline{0.0151}&\underline{0.0820}&\textbf{0.0147}&0.1371 &	0.0364 & 0.1488 &0.0418 
\\

& & & &$\pm 0.0021  $&$\pm 0.0005 $&$\pm 0.0004$&$\pm 0.0003 $&$\pm 0.0066  $&$\pm 0.0017$&$\pm 0.0066 $&$	\pm 0.0031   $&$\pm 0.0062  $&$\pm 0.0034$\\

\midrule
\multirow{6}{*}{Exchange} 
&\multirow{6}{*}{96}  &\multirow{6}{*}{8}&\multirow{2}{*}{4} &\textbf{0.0785}   &\textbf{0.0136} &0.0790 &0.0140 
 &\underline{0.0788} 
   &\underline{0.0137} &0.1357 &0.0359 &0.1615   &0.0490\\
   
& & & &$\pm 0.0026  $&$\pm 0.0010 $&$\pm 0.0002$&$\pm 0.0003 $&$\pm 0.0070  $&$\pm 0.0018$&$\pm 0.0089 $&$	\pm 0.0042  $&$\pm 0.0132  $&$\pm 0.0090$\\

& & &\multirow{2}{*}{8}& \textbf{0.0854}&\textbf{0.0157}&0.0860 
&0.0162 &\underline{0.0857} &\underline{0.0158}&0.1393 &0.0375 &0.1526  &0.0438  
\\

& & & &$\pm 0.0018  $&$\pm 0.0007 $&$\pm 0.0011$&$\pm 0.0003 $&$\pm 0.0068  $&$\pm 0.0019$&$\pm 0.0065 $&$	\pm 0.0031   $&$\pm0.0055  $&$\pm 0.0028$\\

& & &\multirow{2}{*}{10}& \textbf{0.0886}&\textbf{0.0165}&0.0893 &0.0172&\underline{0.0888}&\underline{0.0168}&0.1401 &0.0377 &0.1502 &0.0430
\\

& & & &$\pm 0.0016  $&$\pm 0.0008 $&$\pm 0.0019$&$\pm 0.0005 $&$\pm 0.0022  $&$\pm 0.0008$&$\pm 0.0109 $&$	\pm 0.0054   $&$\pm 0.0042  $&$\pm 0.0019$\\

\midrule
\multirow{6}{*}{Exchange} 
&\multirow{6}{*}{96}  &\multirow{6}{*}{10}&\multirow{2}{*}{4} &\textbf{0.0858}&\textbf{0.0157}&0.0864&0.0162&\underline{0.0862}&\underline{0.0158}&0.1384 &0.0368 &0.1679 &0.0524\\

& & & &$\pm 0.0024 $&$\pm 0.0007 $&$\pm 0.0018$&$\pm 0.0011 $&$\pm 0.0057  $&$\pm 0.0015$&$\pm 0.0027 $&$	\pm 0.0013   $&$\pm 0.0030  $&$\pm 0.0017$\\

& & &\multirow{2}{*}{8}& \textbf{0.0921}&\underline{0.0179}&0.0928&0.0183&\underline{0.0923}&\textbf{0.0178}&0.1426 &0.0389 &0.1536&0.0450
\\

& & & &$\pm 0.0027  $&$\pm 0.0008 $&$\pm 0.0006$&$\pm 0.0005 $&$\pm 0.0067 $&$\pm 0.0020$&$\pm 0.0104 $&$	\pm 0.0052   $&$\pm 0.0038  $&$\pm 0.0016$\\

& & &\multirow{2}{*}{10}&\textbf{0.0949}&\textbf{0.0186} &0.0959&0.0194&\underline{0.0953}&\underline{0.0188}&0.1450 &0.0396  &0.1581 &0.0459 \\

& & & &$\pm 0.0016  $&$\pm 0.0005 $&$\pm 0.0010$&$\pm 0.0006 $&$\pm 0.0077  $&$\pm 0.0024$&$\pm 0.0122 $&$	\pm 0.0063   $&$\pm 0.0058  $&$\pm 0.0033$\\

                             \bottomrule
\end{tabular}
}
\end{table}
}

{\small
\begin{table}
\caption{The results of flexible prediction with different lengths}
\label{flexible_prediction1}
\centering
\resizebox{0.95\textwidth}{!}{
\begin{tabular}{p{12mm}p{6mm}p{6mm}p{6mm}p{8mm}p{8mm}p{8mm}p{8mm}p{8mm}p{8mm}p{8mm}p{8mm}p{8mm}p{8mm}p{8mm}p{8mm}}
\toprule
\multirow{2}{*}{Model}    & \multicolumn{2}{c}{Configs}    & \multicolumn{2}{c}{D2Vformer} & 
\multicolumn{2}{c}{D2Vformer-FT}   &\multicolumn{2}{c}{PatchTST} & \multicolumn{2}{c}{DLinear} & \multicolumn{2}{c}{Fedformer} & \multicolumn{2}{c}{Autoformer} \\
\cmidrule(lr){2-3}
\cmidrule(lr){4-5} \cmidrule(lr){6-7} \cmidrule(lr){8-9} \cmidrule(lr){10-11}  \cmidrule(lr){12-13}\cmidrule(lr){14-15}
                          & \multicolumn{1}{l}{Seq} & $\text{Pred}^*$ & MAE           & MSE           & MAE           & MSE     & MAE           & MSE     & MAE          & MSE          & MAE           & MSE           & MAE            & MSE           \\
                          \midrule
\multirow{4}{*}{ILI}  & \multirow{4}{*}{36}     & 24   & \textbf{0.6345}   & \textbf{1.0152} &  -    &  -       & \underline{0.7149}        & 1.3065       & 0.8433      & 1.4334      & 0.7531       & \underline{1.1744}        & 1.0813       & 1.8063       \\
                          &                         & 12   &0.7427 	&1.3031   &\underline{0.6545} & \underline{1.1300}        & \textbf{0.5677}         & \textbf{0.9098}       & 0.7334        & 1.1384       &0.7110         & 1.1551        & 0.9585          & 1.8284         \\
                          &                         & 20   & 0.7666
         & 1.4066    & \underline{0.7544}  & 1.3426     & \textbf{0.6736 }       & \textbf{1.2869}       & 0.8160       & \underline{1.3271}       & 0.8368        & 1.4274        & 1.0012         & 1.9607        \\
                          &                         & 30   & 0.8601 &	1.7267     &\underline{0.8560} & 1.7074     & \textbf{0.7893 }       & 1.8035       & 0.8663      & \textbf{1.4759}       & 0.8888        & \underline{1.5750}        & 1.0378         & 2.1166        \\
                          \midrule
\multirow{4}{*}{Exchange} & \multirow{4}{*}{96}     & 96   & \textbf{0.1590}        & \textbf{0.0509}   & -& -    & 0.1624        & 0.0540       & \underline{0.1602}       & \underline{0.0521}       & 0.2177        & 0.0886        & 0.2161        & 0.0872        \\
                          &                         & 48   & 0.1171         & 0.0278     &0.1127 & \underline{0.0264}   &\underline{0.1119}       & 0.0270       & \textbf{0.1106}       & \textbf{0.0257 }      & 0.1563        & 0.0467        & 0.1841         & 0.0638        \\
                          &                         & 72   & 0.1406 	&0.0398     &0.1382  &\underline{0.0387}     & \underline{0.1364}        & 0.0391       & \textbf{0.1347}       & \textbf{0.0370}       & 0.1773        & 0.0590        & 0.2065         & 0.0798        \\
                          &                         & 108  &0.1731 	&0.0591   &0.1724 & \underline{0.0579}      & \underline{0.1681}        & 0.0582       & \textbf{0.1661}       & \textbf{0.0539}       & 0.2346        & 0.1046        & 0.2665         & 0.1298        \\
                          \midrule
\end{tabular}
}

\end{table}
}

{\small
\begin{table}[tb]
\centering
\footnotesize
\caption{The results of interval-based flexible prediction}
\label{flexible_prediction2}
\resizebox{0.95\textwidth}{!}{
\begin{tabular}{p{10mm}p{6mm}p{6mm}p{6mm}p{8mm}p{8mm}p{8mm}p{8mm}p{8mm}p{8mm}p{8mm}p{8mm}p{8mm}p{8mm}p{8mm}p{8mm}}
\toprule
\multirow{2}{*}{model}    & \multicolumn{3}{c}{Configs}                                                  & \multicolumn{2}{c}{D2Vformer} & \multicolumn{2}{c}{D2Vformer-FT} &\multicolumn{2}{c}{PatchTST} & \multicolumn{2}{c}{DLinear} & \multicolumn{2}{c}{Fedformer} & \multicolumn{2}{c}{Autoformer} \\
\cmidrule(lr){2-4}
\cmidrule(lr){5-6} \cmidrule(lr){7-8} \cmidrule(lr){9-10} \cmidrule(lr){11-12}  \cmidrule(lr){13-14} \cmidrule(lr){15-16}
                          & \multicolumn{1}{l}{Seq} & \multicolumn{1}{l}{Pred} & \multicolumn{1}{l}{Gap} & MAE           & MSE           & MAE           & MSE  & MAE           & MSE        & MAE          & MSE          & MAE           & MSE           & MAE            & MSE           \\
                          \midrule
\multirow{4}{*}{ILI}  & \multirow{4}{*}{36}     & \multirow{4}{*}{24}      & 2                       & 0.8064 &	1.4661    &\underline{0.8057} & \textbf{1.4598}  & \textbf{0.7406}        & 1.4818       & 0.8772       & \underline{1.4778}       & 0.8878        & 1.6014        & 1.0478         & 1.9957        \\
                          &                         &                          & 4                       & 0.8479 &	1.6245  &\underline{0.8486} & \underline{1.5984}    & \textbf{0.8082}        & 1.6227       & 0.9171       & \textbf{1.5750}       & 0.9058        & 1.6332        & 1.0497         & 2.0477        \\
                          &                         &                          & 6                       &0.8755	&1.6800 &\textbf{0.8672}
                        &  \textbf{1.6259}      & \underline{0.8721}        & 1.8544       & 0.9457       & \underline{1.6667}       & 0.9905        & 1.8251        & 1.0596         & 2.1308        \\
                          &                         &                          & 8                       &0.9492	&2.0675  &0.9081 & 1.8887    &\textbf{0.6967}        & \textbf{1.0693}       & 0.8958       & \underline{1.5102}       & \underline{0.8647}        & 1.5702        & 1.0317         & 2.2175        \\
                          \midrule
\multirow{4}{*}{Exchange} & \multirow{4}{*}{96}     & \multirow{4}{*}{96}      & 2                       & 0.1671& 	0.0547   &0.1663 & \underline{0.0542}    & \underline{0.1651}        & 0.0545       & \textbf{0.1604}       &\textbf{0.0500}       & 0.2103        & 0.0817        & 0.2345         & 0.1027        \\
                          &                         &                          & 4                       & 0.1713 	&0.0568 &0.1698 &\underline{0.0560} 
      & \underline{0.1681}        & 0.0568       & \textbf{0.1643}       & \textbf{0.0519}       & 0.2299        & 0.0998        & 0.2341         & 0.1034        \\
                          &                         &                          & 6                       & 0.1759	&0.0594   & \underline{0.1731}&  \underline{0.0577}   & 0.1741        & 0.0603       & \textbf{0.1684}       & \textbf{0.0538}       & 0.2075        & 0.0800        & 0.2353         & 0.1022        \\
                          &                         &                          & 8                       & 0.1783	&0.0612  & 0.1783&  0.0609 & \underline{0.1758}        & \underline{0.0598}       & \textbf{0.1716}       & \textbf{0.0556}       & 0.2377        & 0.1027        & 0.2377         & 0.1027        \\
                          \midrule

\end{tabular}
}
\end{table}
}

\subsection{Additional Experiments of Flexible Prediction}\label{flexible_prediction}

In Table \ref{skip_exp} of the original paper, we present the performance of various methods on short-term interval-based time series prediction tasks. In the following, we first provide the complete results of this experiment, as shown in Table \ref{deviation}.
Compared to Table \ref{skip_exp}, the main difference is the addition of standard deviation statistics for five repeated experiments of each model. 
From the experimental results, we can see that the standard deviations are relatively small, indicating that the models are minimally affected by random parameters during training and that the outputs are stable. 
Therefore, in other tables in the original paper and appendix, we do not need to present the corresponding standard deviations.

Next, to further reveal the performance of D2Vformer and other baseline models in flexible prediction scenarios, we conducted two additional experiments on tasks with longer prediction lengths: (1) adjacent time series prediction with varying prediction lengths and (2) interval-based time series prediction with varying interval lengths.

\textbf{Adjacent time series prediction with varying prediction lengths:} 
This experiment is also conducted using the ILI and Exchange datasets. 
In the experiment, the D2Vformer undergoes training only once and directly provides predictions of different lengths during the inference phase. 
Other comparison models, however, require retraining and redeployment when the prediction length changes.

For the Exchange dataset, D2Vformer has an input sequence length $\text{Seq}=96$ and a predicted sequence length $\text{Pred}^*=96$ during the training phase. During the testing phase, the input sequence length remains unchanged, while the predicted sequence length $\text{Pred}^*$ changes to $\left\{48, 72, 108\right\}$, and D2Vformer directly outputs the respective predictions.
For the ILI dataset, D2Vformer has an input sequence length $\text{Seq}=36$ and a predicted sequence length $\text{Pred}^*=24$ during the training phase. During the testing phase, when the predicted sequence length $\text{Pred}^*$ is modified to $\left\{12, 20, 30\right\}$, D2Vformer also directly outputs the predictions. Table \ref{flexible_prediction1} shows the prediction errors of D2Vformer and the baseline models in the above scenarios.

It is evident that for medium-term or long-term adjacent flexible prediction tasks, D2Vformer still outperforms some models that have been retrained and redeployed. However, D2Vformer's advantage is less significant than in short-term flexible prediction tasks. This indicates that as the prediction sequence length increases, D2Vformer's adaptability in flexible prediction tasks with different lengths gradually decreases.

In fact, the structure of the D2Vformer model is independent of the length of the prediction time series. Therefore, we can directly perform a small amount of fine-tuning on the original model to meet the requirements of new scenarios. This approach not only further improves the model's performance but also incurs minimal computational costs, as we only need to train a few epochs under the new demands. The performance of D2Vformer after fine-tuning (D2Vformer-FT) is also shown in Table \ref{flexible_prediction1}.

\textbf{Interval-based time series prediction with varying interval lengths:} This experiment is also conducted using the ILI and Exchange datasets. The primary goal is to evaluate the impact on model prediction performance when there is an interval between the input sequence and the predicted sequence, and the interval length 'Gap' changes.
In the experiment, D2Vformer is trained only once and directly provides predictions for sequences with different intervals during the inference phase. In contrast, other comparison models need to be modified when the interval length 'Gap' changes so that their prediction length equals the actual prediction length 'Pred' + 'Gap', after which they are retrained and redeployed.

For the Exchange dataset, D2Vformer has an input sequence length 'Seq' of 96 and a predicted sequence length 'Pred' of 96 during the training phase. During the testing phase, the input sequence length and the predicted sequence length remain unchanged, but the interval length between them successively changes to $\left\{2, 4, 6, 8\right\}$. For the ILI dataset, D2Vformer has an input sequence length 'Seq' of 36 and a predicted sequence length 'Pred' of 24 during the training phase. During the testing phase, the input sequence length and the predicted sequence length remain unchanged, but the interval length between them is successively modified to $\left\{2, 4, 6, 8\right\}$.

Table \ref{flexible_prediction2} shows the prediction results of D2Vformer and the baseline models in this scenario. Compared to Table \ref{skip_exp}, the main difference is the increased length of the predicted sequences. The results in Table \ref{flexible_prediction2} further demonstrate that D2Vformer’s advantage in medium- to long-term flexible prediction tasks is not as significant as in short-term tasks. Additionally, we observe that as the interval length increases, D2Vformer’s adaptability in flexible prediction tasks decreases. However, it is also evident that the direct predictions from D2Vformer are still superior to some models that require retraining and redeployment.
Furthermore, in flexible prediction scenarios with intervals, when the interval length 'Gap' changes, we can also use only a few rounds of fine-tuning to improve the performance of D2Vformer. The corresponding results are named D2Vformer-FT in Table \ref{flexible_prediction2}.

\subsection{Additional Experiments for Effectiveness Analysis}\label{Effectiveness1}

In this section, we conduct supplementary experiments to evaluate the effectiveness of D2Vformer and other methods in various flexible prediction scenarios. The results are presented in Table \ref{tab:flexible_prediction_complexity_interval_based_prediction}, with the left table displaying the outcomes of prediction with varying prediction lengths, and the right table exhibiting the results of prediction with varying interval lengths.
In a variety of flexible prediction scenarios, our D2Vformer demonstrates lower model complexity, as well as reduced training and inference times compared to most of the comparative methods. What's even more notable is that, in dynamically changing scenarios, while all other methods necessitate retraining and redeployment of models to accommodate the updated scenario requirements, D2Vformer does not.

\begin{table}[h]
    \newcommand{\tabincell}[2]{\begin{tabular}{@{}#1@{}}#2\end{tabular}}
    \footnotesize
    \caption{Model complexity evaluation in prediction tasks with varying prediction lengths and interval lengths. 'Numb. of Param.' refers to the number of learnable parameters in the model, 'Inference' indicates the average time the model takes to make predictions for a batch of data, and 'Training' denotes the average time taken to train one epoch of the model.}
\label{tab:flexible_prediction_complexity_interval_based_prediction}
    \begin{subtable}{0.47\textwidth}
        \centering
        \resizebox{1\textwidth}{!}{
            \begin{tabular}{ccllll}
            \toprule
            \multirow{2}{*}{Model}      & \multicolumn{2}{c}{Configs}    & \multirow{2}{*}{\tabincell{c}{Numb. of\\ Param.}}           & \multirow{2}{*}{\tabincell{c}{Training \\(s/epoch)}} & \multirow{2}{*}{\tabincell{c}{Inference \\(ms/batch)}}  \\ \cmidrule{2-3}
                                        & \multicolumn{1}{l}{Seq} & $\text{Pred}^*$ & \multicolumn{1}{c}{}                       & \multicolumn{1}{c}{}                       & \multicolumn{1}{c}{}                           \\
                                        
                                        \midrule
            \multirow{4}{*}{D2Vformer}  & \multirow{4}{*}{96}   & 96   & 146038                                     & 2.4686                                   & 7.4941                                       \\
            &     & 48   & -                                          & -                                          & 7.1410                                       \\
                                        &                         & 72   & -                                          & -                                          & 7.2870                                       \\
                                                              
                                        &                         & 108  & -                                          & -                                          & 7.5000                                         \\
                                        \midrule
            \multirow{4}{*}{PatchTST}   & \multirow{4}{*}{96}   & 96   & 6310336                                    & 6.1478                                   & 5.9440                                       \\
            & & 48   & 5589586                                    & 6.0946                                   & 5.7612                                       \\
                                        &                         & 72   & 6036490                                    & 6.0840                                   & 5.8713                                       \\
                                                               
                                        &                         & 108  & 6856846                                    & 6.2033                                   & 6.1053                                       \\
                                        \midrule
            \multirow{4}{*}{DLinear}    & \multirow{4}{*}{96}  & 96   & 19012                                      & 0.6962                                   & 3.6424                                        \\
            & & 48   & 9312                                       & 0.6652                                   & 3.2410                                       \\
                                        &                         & 72   & 13968                                      & 0.6712                                   & 3.4232                                       \\
                                                               
                                        &                         & 108  & 20952                                      & 0.7209                                   & 4.5437                                       \\
                                        \midrule
            \multirow{4}{*}{Fedformer}  & \multirow{4}{*}{96}  & 96   & 22589454                                   & 29.7278                                   & 79.4551                                       \\ 
            &   & 48   & 19815054                                   & 27.8697                                   & 74.3881                                       \\
                                        &                         & 72   & 21895854                                   & 28.2480                                   & 77.2284                                       \\
                                                                
                                        &                         & 108  & 22589454                                   & 30.5217                                   & 85.7943                                        \\
                                        \midrule
            \multirow{4}{*}{Autoformer} & \multirow{4}{*}{96}   & 96   & 7330252                                    & 13.3049                                   & 78.6800                                        \\
            &  & 48   & 7330252                                    & 11.4382                                   & 70.6933                                       \\
                                        &                         & 72   & 7330252                                    & 12.3086                                   & 75.0954                                       \\
                                                                 
                                        &                         & 108  & 7330252                                    & 15.3784                          &         108.6011 \\
                                        \bottomrule
            \end{tabular}}
    \caption{Prediction with varying prediction lengths}
    \label{tab:flexible_prediction_complexity_different_lengths}
    \end{subtable}
    \begin{subtable}{0.51\textwidth}
        \centering
        \resizebox{1\textwidth}{!}{
            \begin{tabular}{cccllll}
            \toprule
            \multirow{2}{*}{Model}      & \multicolumn{3}{c}{Configs}                              & \multirow{2}{*}{\tabincell{c}{Numb. of\\ Param.}}              & \multirow{2}{*}{\tabincell{c}{Training \\(s/epoch)}}  & \multirow{2}{*}{\tabincell{c}{Inference \\(ms/batch)}}\\
            \cmidrule(lr){2-4}
                                        & \multicolumn{1}{l}{Seq} & \multicolumn{1}{l}{Pred} & Gap & \multicolumn{1}{c}{}                       & \multicolumn{1}{c}{}                       & \multicolumn{1}{c}{}                           \\
                                        \midrule
            \multirow{4}{*}{D2Vformer}  & \multirow{4}{*}{96}     & \multirow{4}{*}{96}      & 2   & 146038                                     & 2.3630                                   & 7.4871                                       \\
                                        &                         &                          & 4   & -                                          & -                                          & 7.4879                                       \\
                                        &                         &                          & 6   & -                                          & -                                          & 7.4834                                       \\
                                        &                         &                          & 8   & -                                          & -                                          & 7.4953                                       \\
                                        \midrule
            \multirow{4}{*}{PatchTST}   & \multirow{4}{*}{96}     & \multirow{4}{*}{96}      & 2   & 6607156                                    & 6.1527                                   & 5.9710                                       \\
                                        &                         &                          & 4   & 6656118                                    & 6.1835                                   & 6.0462                                       \\
                                        &                         &                          & 6   & 6705080                                    & 6.1993                                  & 6.0491                                       \\
                                        &                         &                          & 8   & 6754042                                    & 6.2028                                   & 6.1051                                       \\
                                        \midrule
            \multirow{4}{*}{DLinear}    & \multirow{4}{*}{96}     & \multirow{4}{*}{96}      & 2   & 19022                                      & 0.7025                                   & 3.8092                                       \\
                                        &                         &                          & 4   & 19400                                      & 0.7059                                  & 4.0271                                       \\
                                        &                         &                          & 6   & 19788                                      & 0.7107                                   & 4.1021                                       \\
                                        &                         &                          & 8   & 20176                                      & 0.7199                                   & 4.5381                                       \\
                                        \midrule
            \multirow{4}{*}{Fedformer}  & \multirow{4}{*}{96}     & \multirow{4}{*}{96}      & 2   & 22589454                                   & 29.7664                                   & 82.8762                                       \\
                                        &                         &                          & 4   & 22589454                                   & 29.8867                                   & 84.7731                                       \\
                                        &                         &                          & 6   & 22589454                                   & 30.0998                                   & 85.0564                                       \\
                                        &                         &                          & 8   & 22589454                                   & 30.5013                                   & 85.7214                                       \\
                                        \midrule
            \multirow{4}{*}{Autoformer} & \multirow{4}{*}{96}     & \multirow{4}{*}{96}      & 2   & 7330252                                    & 13.4812                                  & 78.9544                                       \\
                                        &                         &                          & 4   & 7330252                                    & 13.8899                                   & 85.7951                                       \\
                                        &                         &                          & 6   & 7330252                                    & 14.5899                                   & 92.7951                                       \\
                                        &                         &                          & 8   & 7330252                                    & 15.0245                                   & 108.1761 \\
                                        \bottomrule
            \end{tabular}}
        \caption{Prediction with varying interval lengths}

    \end{subtable}

\end{table}

\end{document}